\documentclass[10pt,twocolumn,letterpaper]{article}

\usepackage[pagenumbers]{wacv} 
\usepackage{graphicx}
\usepackage{amsmath}
\usepackage{amssymb}
\usepackage{booktabs}
\usepackage{multirow}
\usepackage{amssymb}
\usepackage{xcolor}
\usepackage{float}

\usepackage[pagebackref,breaklinks,colorlinks]{hyperref}

\usepackage[capitalize]{cleveref}
\crefname{section}{Sec.}{Secs.}
\Crefname{section}{Section}{Sections}
\Crefname{table}{Table}{Tables}
\crefname{table}{Tab.}{Tabs.}

\def\wacvPaperID{2377} 
\def\confName{WACV}
\def\confYear{2024}

\begin{document}

\title{P-Age: Pexels Dataset for Robust Spatio-Temporal Apparent Age Classification}

\author{Abid Ali*\\
Inria, France\\
Universit\'e Cote d'Azur, France\\
{\tt\small name.surname@inria.fr}
\and
Ashish Marisetty*\\
IIIT Naya Raipur, India\\
{\tt \small mashish1305@gmail.com}
\and
Francois Bremond\\
Inria, France\\
Universit\'e Cote d'Azur, France\\
{\tt\small name.surname@inria.fr}}
\maketitle

\begin{abstract}
Age estimation is a challenging task that has numerous applications. In this paper, we propose a new direction for age classification that utilizes a video-based model to address challenges such as occlusions, low-resolution, and lighting conditions. To address these challenges, we propose AgeFormer which utilizes spatio-temporal information on the dynamics of the entire body dominating face-based methods for age classification. Our novel two-stream architecture uses TimeSformer and EfficientNet as backbones, to effectively capture both facial and body dynamics information for efficient and accurate age estimation in videos. Furthermore, to fill the gap in predicting age in real-world situations from videos, we construct a video dataset called Pexels Age (P-Age) for age classification. The proposed method achieves superior results compared to existing face-based age estimation methods and is evaluated in situations where the face is highly occluded, blurred, or masked. The method is also cross-tested on a variety of challenging video datasets such as Charades, Smarthome, and Thumos-14. The code and dataset is available at \url{https://github.com/Ashish013/AgeFormer}.
\let\thefootnote\relax\footnotetext{* authors contributed equally}

\end{abstract}

\section{Introduction}
\label{sec:intro}
Age estimation is a challenging task that has many applications in various domains such as social robotics, video surveillance, business intelligence, social networks, and demography. Typically, the goal of age estimation involves predicting the age of a person by his/her facial appearance, which can be affected by many factors such as resolution, lighting condition, pose, expression, occlusion, and makeup. In recent years, numerous attempts have been made to estimate the age of a person from facial images \cite{liu2015agenet, sheoran2021age, rothe2015dex, zhang2017quantifying, zhu2015study, zhang2019fine, levi2015age, tan2017efficient}. 

Age estimation models can be broadly classified into two categories: regression-based and classification-based. Regression-based models directly output a numerical value for the age, whereas classification-based models assign the image to one of the predefined age groups. Recent advances in age estimation are mainly driven by the development of deep learning techniques, especially Convolutional Neural Networks (CNNs). Several methods adopted 2D CNNs to estimate age from a single-face image. Furthermore, researchers recently combined 2D-CNNs with attention-mechanism \cite{wang_adpf_2021}, adversarial learning \cite{penghui_2019}, domain adaptation \cite{Singh2020}, and multi-task learning techniques \cite{yoo_deep_facial_2018, sym10090385} to improve the robustness of age estimation models. However, estimating the age of a single-face image is still an open challenge due to several factors such as \textbf{occlusions}, \textbf{low-resolution}, and \textbf{lighting conditions}. In addition to that, face-based age estimation models fail in situations where the face is not \textbf{visible} all the time or \textbf{blurred} due to privacy concerns. We argue that a single-face image is not enough to overcome these challenges in real-world situations. Therefore, a new technique based on video that captures the semantics of the entire body is needed to address these challenges. 

3D-CNNs have been widely used in domains such as video analysis \cite{feichtenhofer2020x3d,
 tran2014c3d}, action recognition \cite{wang2016temporal,ali:hal-03447060, Kahatapitiya_2021_CVPR}, and medical image segmentation \cite{NIYAS2022397}, due to their ability to capture spatial and temporal information together. A video-based model can be useful to capture the spatio-temporal features of the frames to predict the age of a person in real-world situations such as surveillance and or medical assessment, where the face is not visible all the time. In addition, modern techniques, such as transformers \cite{vaswani2017attention} can be used to further handle long-range dependencies in an environment where the face is highly occluded. Furthermore, we believe that using only facial features is not enough to predict age. In addition, other characteristics of the body, such as the \textbf{ body dynamics of a person} (face, head, and limbs, etc.) can provide significant cues to estimate the age of the person. Therefore, we propose a video-based age estimation model that captures significant spatio-temporal features of the entire body to improve the age estimation task. We reuse existing video models for a completely new problem (age predictions). Our results on unseen data such as Charades, Smarthome, and Thumos-14 Figure \ref{fig:cross} demonstrate the novelty of utilizing such video models for a completely different problem such as age classification. This opens up new research directions for problems other than age classification, such as gender, ethnic groups, and other demographic classifications. Many works that rely on faces can be extended to video-based models to capture body dynamics for robust recognition.
 
In summary, in this paper, we propose a new direction for age estimation utilizing video-based models to address the above challenges. We introduced a new dataset, \textbf{P-Age}, for video-based age classification. Furthermore, we propose a novel two-stream architecture to improve age prediction in real-world scenarios. Our architecture uses TimeSformer \cite{gberta_2021_ICML} in combination with EfficientNet \cite{tan2019efficientnet} as the backbone. The features of both streams are fused using a multihead-attention module for age classification. 
Some of the main contributions of this paper are as follows:
\begin{itemize}
    \item To the best of our knowledge, we are the first to propose a video-based method for age classification in challenging situations. Our novel \textbf{AgeFormer} architecture achieves superior results compared to existing face-based age estimation methods.
    \item We propose a new dataset called Pexels Age (\textbf{P-Age}), scrapped from Pexels (a royalty-free stock footage website) for age classification into four distinct groups. Additionally, we provide baseline results on this dataset and compare our method with existing face-based methods on the P-Age-Face dataset.
    \item Keeping privacy preservation a priority, we evaluate our method in situations where the face is i) \textbf{extremely occluded}, ii) \textbf{blurred}, and iii) \textbf{masked}. In addition, we validate the efficiency of our method in real-world situations by testing it on a variety of challenging video datasets such as \textit{Charades} \cite{sigurdsson2016hollywood}, \textit{Smarthome} \cite{dai2022toyota}, and \textit{Thumos-14} \cite{IDREES20171}.
\end{itemize}
\begin{figure*}[h!]
\begin{center}
\includegraphics[width=0.7\linewidth]{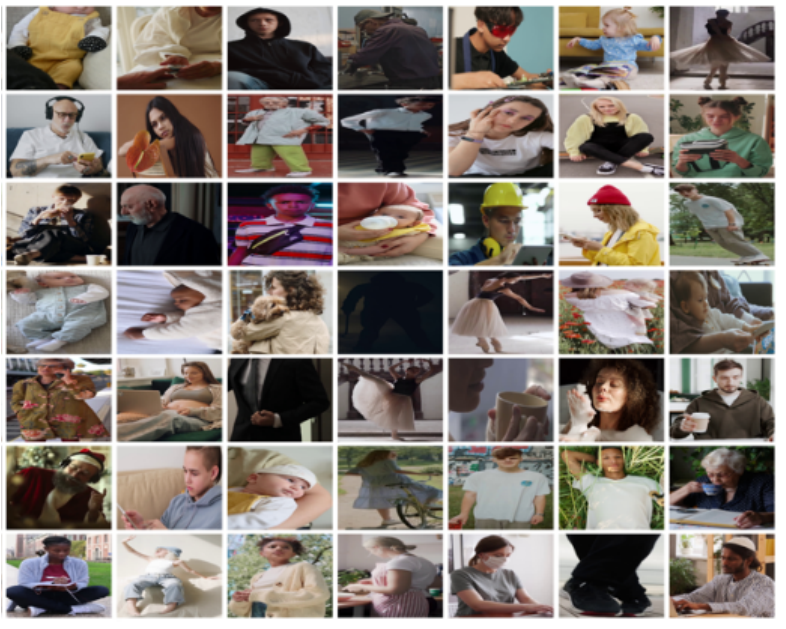}
\end{center}
   \caption{An overview of the P-Age dataset.}
\label{fig:dataset}
\end{figure*}
\section{Related Work}
\textbf{Age Estimation:} Deep learning-based methods use convolutional neural networks (CNNs) to automatically learn features from large-scale face datasets \cite{levi2015age, shin2022moving, 7780901, shen2017learning}. Age-pattern learning methods can be categorized into classification-based and regression-based approaches. Classification-based methods treat age estimation as a multi-class problem, where each class corresponds to an age group or a single year \cite{geng2006learning, levi2015age, kwon1999age}. AGNet \cite{liu2015agenet} uses CNNs network with a softmax layer to classify the images into 8 age groups. Recently, PML \cite{Deng_2021_CVPR} proposes a progressive marginal loss approach for the classification of unconstrained facial age.

Regression-based methods treat age estimation as a continuous problem, where the output is a real-valued age estimate. DEX \cite{rothe2015dex} uses CNNs to learn the mapping of image pixels to age labels and then computes the expected value of the age distribution as the final prediction. OR-CNN \cite{7780901} converts the age estimation problem into an ordinal regression problem, where each neuron in the output layer represents an ordinal number, and the model learns to activate neurons that correspond to the true age or lower. SSR-Net \cite{ijcai2018p150} adopts a multi-stage design that progressively refines age prediction from coarse to fine. More recently, \cite{shin2022moving} proposed a new ordinal regression algorithm called Moving Window Regression (MWR) that introduces the notion of relative rank (rho-rank) and develops global and local relative regressors (rho-regressors) to predict rho-ranks within entire and specific rank ranges, respectively, for age estimation from a single face image.

Furthermore, several datasets have been proposed for age estimation in recent years, including UTKFace \cite{chandaliya2020scdae}, MORPH \cite{ricanek2006morph}, CACD \cite{chen2015face}, FG-NET \cite{panis2016overview}. However, these datasets are image-based mainly focusing on face-crops. Therefore, the methods introduced in the past are limited to face-crops only. In contrast, we propose a video-based model to learn spatial and temporal features from the semantics of the entire body. 

\textbf{Video Classification}: CNNs have shown great success in learning 3D spatio-temporal representations to recognize human actions \cite{carreira2017quo, tran2014c3d, tran2017convnet, hara2017learning}. Two-stream techniques, frequently combined RGB with optical flow \cite{feichtenhofer2016convolutional, simonyan2014two}, to classify videos. SlowFast network \cite{feichtenhofer2019slowfast} has shown that action recognition can be enhanced by mixing representations of various temporal resolutions (i.e. frame rates). More recently, with the introduction of Transformers, numerous systems have improved action recognition by including attention \cite{Arnab_2021_ICCV, Li_2022_CVPR, Wang_2022_CVPR, Dai_2021_WACV, Dai_2022_CVPR}. TimeSformer \cite{gberta_2021_ICML} introduced a divided space-time attention mechanism to capture spatio-temporal dependencies throughout the video. 

To go beyond Video Classification, our architecture utilizes action recognition architectures for age estimation tasks. We combine an action recognition (video) model with a 2D-CNNs method (image) to create a two-stream network for robust age classification.
\begin{table*}[h]
\begin{center}
\begin{tabular}{cccccc}
\hline
\multirow{2}{*}{Age Group} & \multirow{2}{*}{Downloaded Videos} & \multicolumn{1}{c|}{\multirow{2}{*}{Filtered Videos}} & \multicolumn{3}{c}{\# of Frames} \\ \cline{4-6} 
                       &                                    & \multicolumn{1}{c|}{}                                 & Min.      & Avg.      & Max.     \\ \hline
Baby / Toddler         & 1760                               & 401                                                   &  4         &   315        &    1708      \\
Adolescent             & 880                                & 531                                                   &   2        &    393       &     2436     \\
Adult                  & 1048                               & 416                                                   &   18        &    485       &    3523      \\
Elderly                & 880                                & 324                                                   &   2        &     426      &    2106      \\ \hline
Total                  & 4568                               & 1672                                                  &    -       &     -      &     -     \\ \hline
\end{tabular}
\end{center}
\caption{Statistics of the P-Age dataset.}
\label{tab:state}
\end{table*}

\section{P-Age Dataset}
In this section, we discuss the key choices in creating the Pexels Age (P-Age) classification dataset. First, we discuss the data scraping process. Then, we explain the annotation procedure and the quality control checks to refine the annotations.

\subsection{Data Preparation}
The P-Age dataset is sourced from Pexels (a royalty-free stock footage website) \cite{pexel} using search keywords such as \textit{baby, toddler, teen, man, woman, elderly, old, etc.}. Furthermore, our dataset represents different ethnic groups (\textit{African, European, etc.}). An overview of the dataset is given in Figure \ref{fig:dataset}. We downloaded more than 4,500 videos with 27-fps and a high-quality 1080p resolution. The average video has more than 400 frames and 5.7\% of the videos in the P-Age dataset do not have a face appearance. Furthermore, most of the videos in our dataset have occluded faces (for example, a cinematic video of a child playing initially from his/her legs and slowly changing the camera to reveal the face). 

\subsection{Annotation}
The P-Age dataset is divided into four age groups: baby/child, teen, adult, and elderly based on the searched query. For each class, we downloaded at least 880 videos per class and made sure that the gender ratio was balanced by using gender-specific keywords such as \textit{baby-girl, teen-boy, etc.}. To further save time, we annotated the dataset based on the searched query type (child, teen, adult, elderly).
\begin{figure*}[htbp]
\begin{center}
\includegraphics[width=\linewidth]{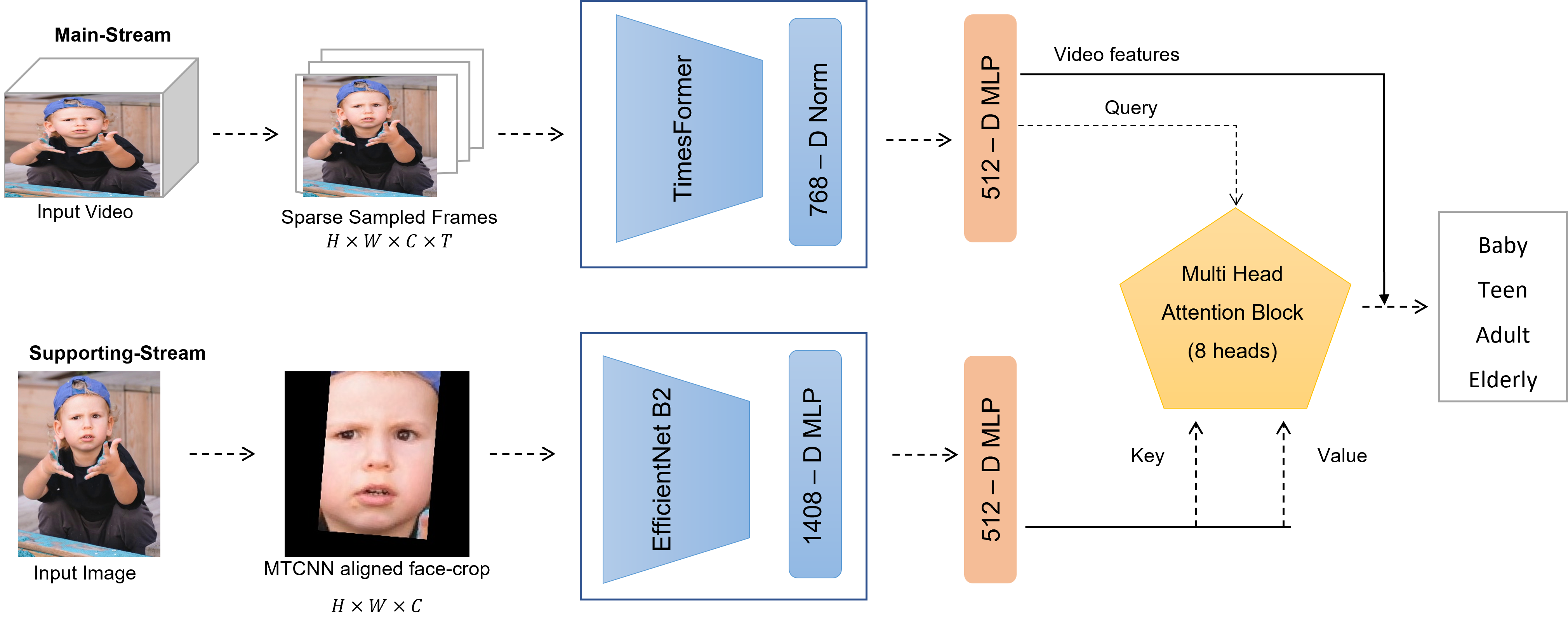}
\end{center}
   \caption{Proposed AgeFormer architecture. The main-stream learns spatio-temporal features of body dynamics, while the supporting-stream provides additional face features to improve age prediction.}
\label{fig:ageformer}
\end{figure*}

\begin{table*}[h]
\begin{center}
\begin{tabular}{ccccc}
\hline
\textbf{}      & \textbf{\# videos} & \textbf{\# images} & \textbf{ occlusions} & \textbf{\# videos having no faces} \\ \hline
MORPH \cite{ricanek2006morph}   & $\times$             &          55.1k            & $\times$                      & $\times$                                  \\
UTKFace \cite{chandaliya2020scdae} & $\times$               &          23.7k            & $\times$                      & $\times$                                  \\
P-Age        &        1672            &       675.5k          &        $\checkmark$                &      95                          \\ \hline
\end{tabular}
\end{center}
\caption{Comparison with age estimation benchmark datasets.}
\label{tab:dataset_comparison}
\end{table*}

\subsection{Quality Control}
The downloaded videos are then cleaned with their brief video filenames to ensure that the video actually belongs to the class to which it is assigned, and videos that do not belong to any class are removed.
Several videos have more than one person. We use existing object detection and tracking tools to filter out videos that have multiple people or no people. We utilize Yolov5 \cite{glenn_jocher_2022_7347926} with OSNet \cite{zhou2019omni} to filter out such videos. Furthermore, three experts visually cross-check the annotation to correct any incorrect videos and remove outliers. After filtering, we have obtained 1672 videos; per-class statistics of the dataset are given in Table \ref{tab:state}. A comparison is noted in Table \ref{tab:dataset_comparison} with existing age estimation datasets.

\subsection{P-Age-Face}
Furthermore, to compare with other age estimation methods, we have created a face-only variant of the P-Age dataset called \textbf{P-Age-Face} by randomly sampling 10 frames in a stride of 4 from each video. Following the general pre-processing convention outlined in \cite{levi2015age}, we extract the faces using MTCNN \cite{zhang2016joint} and then use the eye landmark coordinates as a reference to align the face crops. Since 5.7\% of the videos in our dataset lack faces, the P-Age-Face dataset accounts for 94. 3\% of the size of the P-Age dataset. Upon acceptance, both datasets will be made available to the general public.

\section{AgeFormer}
In this section, we discuss our proposed architecture for robust age classification in videos. Our design is inspired by action recognition networks and how they handle the spatio-temporal information to recognize an action type. In contrast to action recognition methods, where the model learns gestures and body movements over time, we learn the dynamics of the entire body (\textit{face, head, height and width of limbs, etc.}) over time. Here, the time information helps the model to learn the appearance of the person at different frames in a video snippet. For example, in the starting frames of a video, only the lower body is visible to the camera, but after a few seconds, the entire person appears (face or upper body appearance). Thus, the full appearance information is compatible with partially visible frames. The overall architecture is shown in Figure \ref{fig:ageformer}. \\


\subsection{Video-stream}
The main-stream is a video transformer network that uses TimeSformer \cite{gberta_2021_ICML} as the backbone. TimeSformer \cite{gberta_2021_ICML} is a convolution-free method to classify videos that relies solely on individual attention in space and time. By allowing spatio-temporal features to learn directly from a series of frame-level patches, it applies the traditional Transformer architecture to videos. The main-stream takes an input clip $X \in \mathbb{R}^{H \times W \times 3 \times T}$ having $T$ frames of size $H \times W$. The input frames are decomposed into $N$ patches of size $P \times P$. The patches are flattened into vectors $X_{(p, t)} \mathbb{R}^{3P^2}$, where $p = 1,...,N$ denotes spatial locations and $t = 1,...,T$ defines the frame index. The readers can refer to the TimeSformer \cite{gberta_2021_ICML} paper for more details. The main-stream captures the spatio-temporal dynamics of the entire body.

\subsection{Image-Stream}
The supporting stream is a 2D-CNN network that uses EfficientNet B2 \cite{tan2019efficientnet} as the backbone. EfficientNet is a robust and high-performing image classification network that has fewer parameters and is less computationally expensive, making it the perfect choice to complement the parameter-heavy video backbone. The supporting stream takes the face-crops extracted using MTCNN \cite{zhang2016joint}, aligned by the eye landmarks from the P-Age-Face dataset as input $X \in \mathbb{R}^{H \times W \times 3}$, where $H$ and $W$ represent height and width. This stream learns additional facial features that are known to be good indicators to further improve age estimation. In addition, a zero matrix is provided as input to the supporting stream when there is no face information available. More details of EfficientNet can be found in \cite{tan2017efficient}. 

\subsection{Attention Mechanism}
A 768 feature vector from the main-stream, and a 1408 feature vector from supporting-stream are passed through an MLP layer obtaining 512 dimensional vectors for each stream after applying dropout and layer-normalization. The two streams are concatenated using a multi-head-attention (MHA) module having 8 attention heads. Query $Q$ is obtained from the main-stream attended with the Key and Value \{$K, V$\} pair acquired from the supporting-stream. A skip connection is added between the main-stream output and the MHA module output and passes through a linear layer with a softmax activation function to predict the age class. This architecture allows our model to effectively capture both facial and body dynamics information in the input for efficient and accurate age estimation using videos.


\section{Experiments}
Our experiments are categorized into four folds. First, we compare different video-based methods on the P-Age dataset, providing baseline results. Second, we compare our AgeFormer with existing face-based SOTA methods on the P-Age-Face dataset. Third, we evaluate the quality of our method in robust situations such as privacy preservation (occluded, blurred, or blacked-out faces). Finally, we cross-test AgeFormer on different challenging video datasets such as THUMOS-14 \cite{IDREES20171}, Smarthome \cite{dai2022toyota} and Charades \cite{sigurdsson2016hollywood}. Further details such as analysis of low-resolution, the effect of different sampling-rates, and situations where our model did not perform well are provided in the Supplementary Materials.
\subsection{Experimental Details}
\label{sec:experimental materials}
All our models are implemented using the PyTorch library. We leverage the pre-trained weights from Kinetics-400 and ImageNet to initialize the models for the video and image streams, respectively. The main-stream receives 32 (temporal length) 224 $\times$ 224 resolution frames sampled at a stride of 4, while the supporting-stream takes in a 288 $\times$ 288 resolution face crop. The proposed architecture and video baselines are trained for 25 epochs with a batch size of 16 using an AdamW optimizer with a learning rate of $3e^{-5}$ and an exponential learning rate scheduler with a gamma of 0.9 using a cross-entropy loss. On the other hand, the face classifier benchmark training is conducted for 200 epochs, with a batch size of 512 and a learning rate of $2e^{-3}$. These hyper-parameter settings have been configured to ensure the convergence of all models. Furthermore, all of the performance analysis experiments involve only changing one parameter in the study (resolution, face-blur, stride) at a time, while the rest of the model remains unchanged and in inference mode.

Moreover, The P-Age dataset is split in a stratified way into train, validation, and test sets that have a proportion of 76\%, 12\%, and 12\%, respectively.
\begin{table}[h]
\begin{center}
\begin{tabular}{ccccc}
\hline
Method             & Acc.             & Precision/Recall & F1-Score \\ \hline
X3D \cite{feichtenhofer2020x3d}         & 77.61\%          & 0.80 / 0.76      & 0.77 \\
Slow-Fast \cite{feichtenhofer2019slowfast}   & 78.11 \%          & 0.79 / 0.78      & 0.78    \\
I3D \cite{carreira2017quo}         & 81.59\%          & 0.82 / 0.82      & 0.82  \\
MViT \cite{fan2021multiscale}        & 86.00\%          & 0.86 / 0.87      & 0.86   \\ \hline
TimeSformer \cite{gberta_2021_ICML} & 87.56\%          & 0.88 / 0.88      & 0.88   \\
\textbf{Proposed} & \textbf{89.55\%}          & \textbf{0.90 / 0.90}      & \textbf{0.90}  \\ 
\hline
\end{tabular}
\end{center}
\caption{Baseline results of different video-based methods on the P-Age dataset.}
\label{tab:baseline}
\end{table}

\subsection{Results and Discussion}
We divide our experiments into two categories. First, we create a baseline for video-based age classification. We evaluate different action classification architectures for the age-estimation task on the P-Age dataset. We compare both 3D-CNNs and transformer-based methods for this task, as shown in Table \ref{tab:baseline}. Our AgeFormer achieves SOTA results. Furthermore, the results indicate that video-based methods, specifically transformer architectures, are good enough to estimate the age of an individual. Adding an additional face-branch as in our AgeFormer, can further improve accuracy.

\begin{table}[h]
\begin{center}
\begin{tabular}{cccc}
\hline
Method             & Acc.             & Precision/Recall & F1-Score  \\ \hline
C3AE\cite{zhang2019c3ae}               &37.70\%            &0.21 / 0.30      & 0.21    \\
MWR \cite{shin2022moving}         & 43.39\%          & 0.40 / 0.42      & 0.39      \\
Kim \cite{kim2021generalizing}             & 57.67 \%         & 0.68 / 0.58      & 0.59      \\
ResNet-18 \cite{Nebula4869}        & 61.38 \%         & 0.68 / 0.60      & 0.61      \\
Levi \cite{levi2015age}             & 70.37\%          & 0.70 / 0.69      & 0.69      \\
\hline
EfficientNet B2             & 53.97\%          & 0.66 / 0.55      & 0.52 \\
\textbf{Proposed*} & \textbf{77.25\%}          & \textbf{0.79 / 0.77}      & \textbf{0.77}                \\
\textbf{Proposed} & \textbf{89.42\%}          & \textbf{0.89 / 0.90}      & \textbf{0.89} \\
\hline
\end{tabular}
\end{center}
\caption{Comparison of existing face-classifiers with proposed video-based classifiers on P-Age-Face dataset. * means a video-based model (only main-stream) that utilizes the whole body with blurred faces as input. EfficientNet B2 results are from our supporting-stream (face model).}
\label{tab:comparison}
\end{table}

\begin{figure*}[H]
\begin{center}
\includegraphics[width=\linewidth]{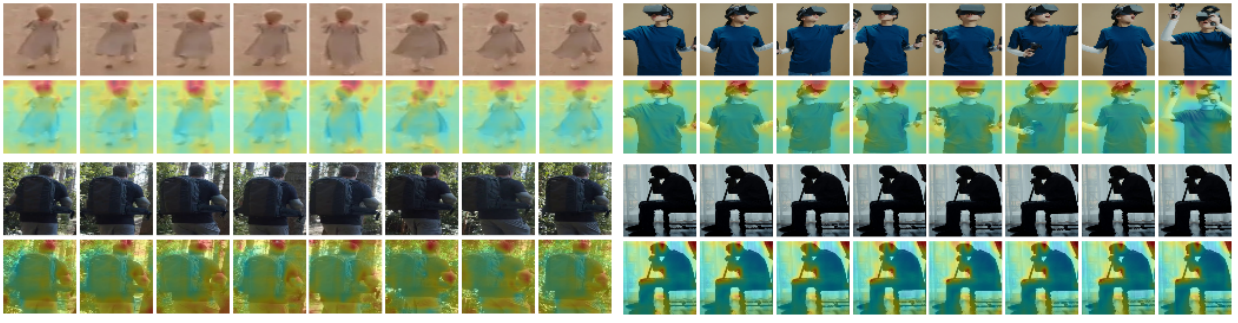}
\end{center}
   \caption{Spatio-temporal attention visualization of AgeFormer on P-Age dataset.}
\label{fig:atten}

\end{figure*}
\begin{figure*}[H]
\begin{center}
\includegraphics[width=\linewidth]{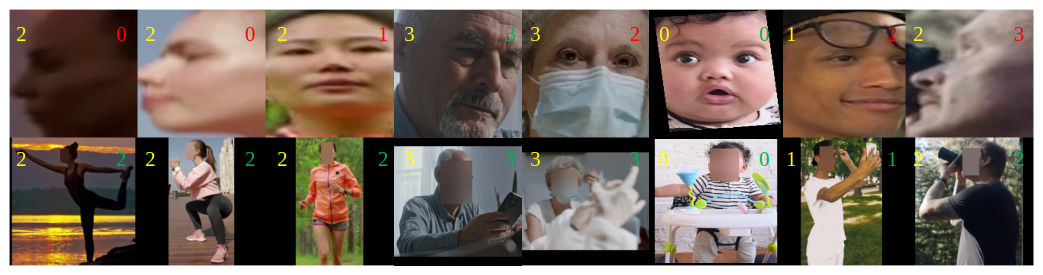}
\end{center}
   \caption{Comparison of SOTA face-based age classifier \cite{levi2015age} with our proposed video-based method. Our method takes a video input with blurred faces.}
\label{fig:privacy}
\end{figure*}
Second, we utilize the Face variant of the P-Age dataset to make use of the SOTA face-based methods for comparison with our AgeFormer. AgeFormer achieves new SOTA results compared to face-based methods on the P-Age dataset. We compare our method with face-based methods in two ways as noted in Table \ref{tab:comparison}. The results indicate that our proposed method is superior to existing methods, even when the faces are blurred. This validates our idea of using space-time dependencies of the entire body. A more qualitative comparison between Ageformer and SOTA method is provided in the Supplementary Materials.

\subsection{Important Cues for Predicting Age}
\begin{figure*}[t]
\begin{center}
\includegraphics[width=\linewidth]{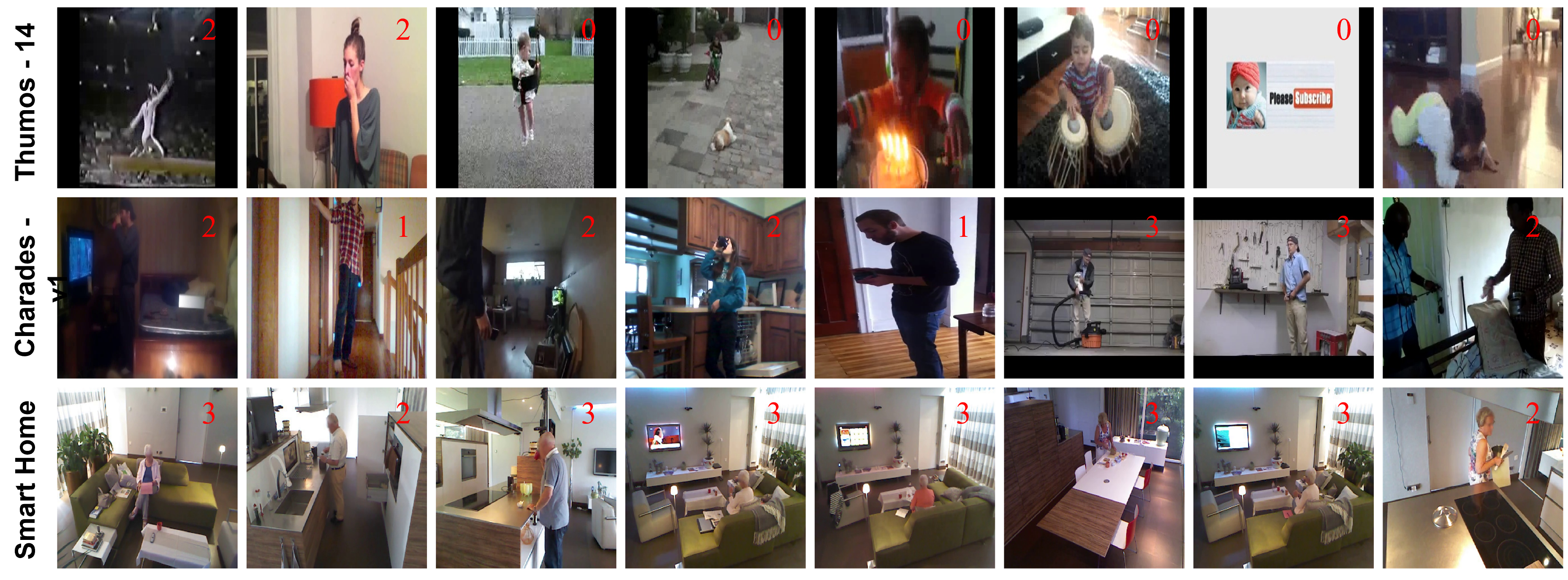}
\end{center}
   \caption{Each row depicts eight randomly selected videos from a specific dataset, the annotated \textcolor{red}{label} is predicted by AgeFormer. We observe that despite having never been trained on them, the model performs reasonably well.}
\label{fig:cross}

\end{figure*}

We visualize the divided space-time attention weights of the video backbone to provide insight into which part of the input is the most important when predicting age. The body of a person (limbs, chest, torso) and its dynamics hold important cues to differentiate one age group from the other. Our model learns these cues as shown in the Figure \ref{fig:atten}. Using the attention roll-out method, we can visualize that the model is attentive to not just the face, but also other body parts, especially in the case where the face is not visible Figure \ref{fig:atten} (second row, first example). Furthermore, the prediction of age from the body dynamics of a person has been validated in unseen videos in Figure \ref{fig:cross} (\textcolor{red}{Smarthome}), where the faces are barely visible, but our model can accurately predict age in such situations.
\begin{table}[h]
\begin{center}
\begin{tabular}{cccc}
\hline
Augmentation        & Acc.              & Pre./Recall     & F1-Score      \\ \hline
face-blur        & 77.61 \% & 0.80 / 0.78 & 0.78 \\
face blacked-out & 76.88\%  & 0.79 / 0.76 & 0.76 \\ \hline
\end{tabular}
\end{center}
\caption{Effectiveness of the proposed approach in the situation of extreme privacy preservation in the P-Age dataset.}
\label{tab:face_comparison}
\end{table}

\begin{figure*}[!ht]
\begin{center}
\includegraphics[width=\linewidth]{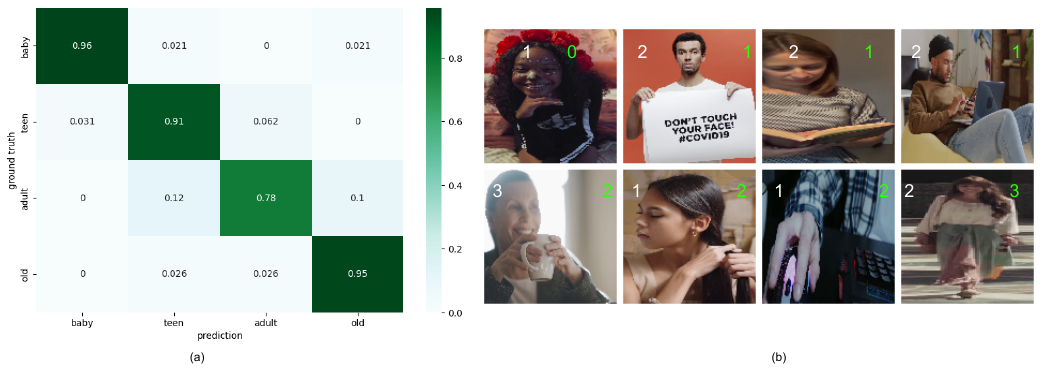}
\end{center}
   \caption{(a) illustrates the confusion matrix of the proposed method and (b) shows bad cases of our proposed method. White color indicates ground-truth, and \textcolor{green}{green} indicates the predicted label.}
\label{fig:cm_bad}
\end{figure*}
\subsection{Privacy Preservation}
In this section, we discuss the privacy preservation aspect of our method. In real-world applications such as activities of daily living, surveillance, and medical evaluations, most of the data are sensitive to privacy concerns, where faces are often obscured, blurred, or blacked-out. Furthermore, in cases such as autism assessment for children \cite{ali:hal-03447060}, the movements of the child cannot be controlled, and therefore their faces are not visible all the time, or, in other words, the upper-body is highly occluded. In such cases, existing face-based methods do not categorize the age of the participants. 

Therefore, our AgeFormer is superior to prior face-based methods by incorporating spatio-temporal features of the entire body. This helps the model classify the age more efficiently, as demonstrated in Figure \ref{fig:privacy}, even in scenarios where the faces are blurred-out. In these experiments, we use only the main-stream of our architecture. We further evaluate our model on two scenarios, i) blurred, and ii) blacked-out faces. The results in Table \ref{tab:face_comparison} validate our idea of using a video-based model with the whole body as input to preserve privacy.

\subsection{Qualitative Cross Dataset Performance}
In this section, we evaluate the generalizability of our proposed AgeFormer. We perform a qualitative cross-dataset performance analysis on real-world video (action recognition) benchmarks such as THUMOS-14 \cite{IDREES20171}, Smarthome \cite{dai2022toyota}, and Charades \cite{sigurdsson2016hollywood} without training on these datasets. These datasets are selected because together they contain people from all age groups with significant variations in facial expressions, poses, lighting conditions, and occlusions. To this extent, we first randomly sample 30 different videos from each dataset and perform person detection and tracking to obtain the various person videos (tracklet). Following that, we use AgeFormer to predict the age class of each person (tracklet). As there are no ground-truth labels available for these datasets, we assign only the predicted labels to each person, as illustrated in Figure \ref{fig:cross}. 

Our results show that, although the model was never trained on low-resolution data, the proposed approach is resilient in most scenarios, except for a few cases where the video quality is severely degraded (Figure \ref{fig:cross} (1,1)), extreme low light conditions (Figure \ref{fig:cross} (2,8)), or the apparent age of the person is uncertain and falls between two categories (Figure \ref{fig:cross} (3,8)). If a facial crop is achievable, the latter issue could be potentially resolved by leveraging the facial feature context of the supporting stream, leading to more consistent predictions and better performance. In general, the proposed two-stream method is an effective approach with broad generalizability by extracting the relevant contextual features from the inputs.

\subsection{Bad Cases}
The proposed method may not perform well in tricky situations, where it is even difficult for humans to predict the age of an individual, as shown in Figure \ref{fig:cm_bad}\textcolor{red}{b}. The confusion matrix in Figure \ref{fig:cm_bad}\textcolor{red}{a} demonstrates that the model has a hard time classifying adults. In such difficult cases, the proposed model mostly confuses the teen and adult classes. Furthermore, the visual appearance of the individuals in the first column of Figure \ref{fig:cm_bad}\textcolor{red}{b} illustrates why the model incorrectly predicts age.
\section{Conclusion}
Age-classification of an individual in real-world situations is a challenging task. In this paper, we propose a new direction to predict the age of an individual in a video. Our novel video-based model named AgeFormer achieves a precise age classification in challenging situations. The proposed architecture utilizes spatio-temporal information of the dynamics of the entire body dominating face-based methods for age classification. Experiments illustrate that video-based models are robust against situations where the face is obscured (privacy preservation) to predict the age of an individual. Additionally, we built the first video dataset (P-Age) for age classification, opening the door for researchers to explore video-based methods for age classification.
\section*{Acknowledgments}
The COFUND BoostUrCAreer program has received funding from the European Union’s Horizon 2020 research and innovation program under the Marie Curie grant agreement No 847581, from the Région SUD Provence-Alpes-Côte d'Azur and IDEX UCAjedi.
\begin{figure}[h!]
\begin{centering}
\includegraphics[width=0.6 \columnwidth]{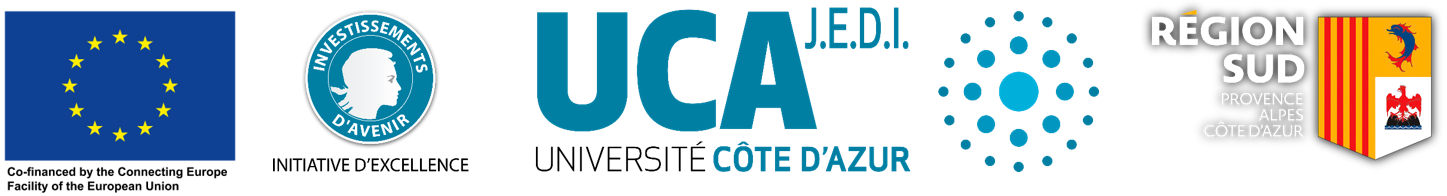}
\end{centering}
\end{figure}
{\small
\bibliographystyle{ieee_fullname}
\bibliography{egbib}
}
\end{document}


\title{Supplementary: P-Age: Pexels Dataset for Robust Spatio-Temporal Apparent Age Classification}

\author{Abid Ali*\\
Inria, France\\
Universit\'e Cote d'Azur, France\\
{\tt\small name.surname@inria.fr}
\and
Ashish Marisetty*\\
IIIT Naya Raipur, India\\
{\tt \small mashish1305@gmail.com}
\and
Francois Bremond\\
Inria, France\\
Universit\'e Cote d'Azur, France\\
{\tt\small name.surname@inria.fr}}
\maketitle

\begin{figure*}[htbp]
\begin{center}
\includegraphics[width=0.9\linewidth]{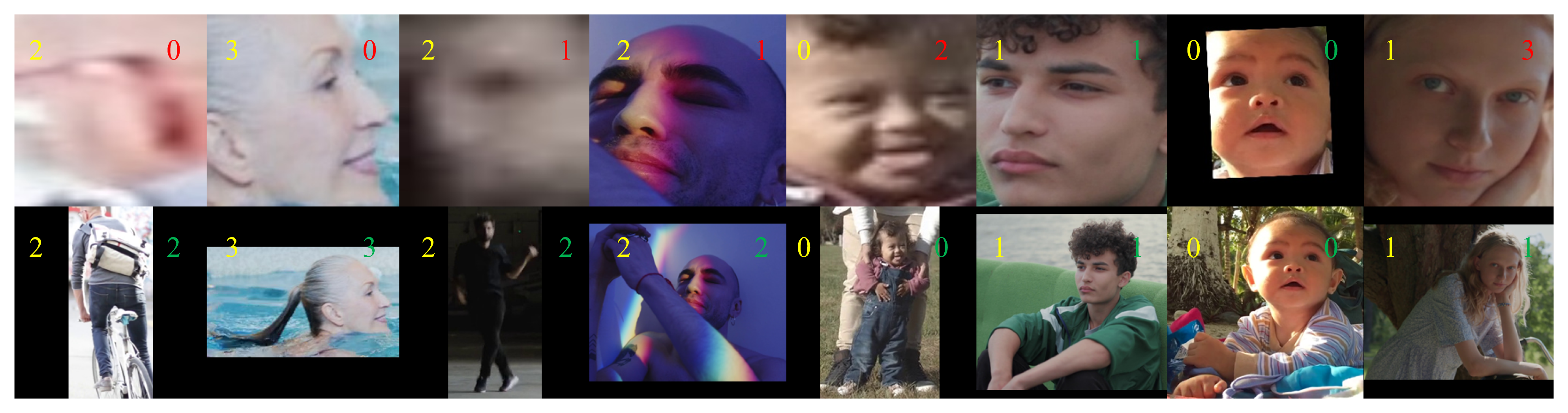}
\end{center}
   \caption{Qualitative comparison of SOTA with proposed method on the P-Age-Face dataset. The top-row are prediction from the best SOTA, while bottom-row are from AgeFormer. The \textcolor{yellow}{yellow} color represent ground-truth, while \textcolor{green}{green} (correct), and \textcolor{red}{red} (incorrect) define predictions from the model.}
\label{fig:qual_comp}
\end{figure*}
\section{Qualitative Results}
In addition to the tabular results in Table \textcolor{red}{4} of the main artcile, we perform a qualitative analysis as shown in Figure \ref{fig:qual_comp} between the best SOTA Levi \cite{levi2015age} age classifier and our AgeFormer on the P-Age-Face dataset. The results indicate that our AgeFormer can accurately predict age in situations such as low-resolution, obscured faces, etc. 

\begin{figure*}[htbp]
\begin{center}
\includegraphics[width=\linewidth]{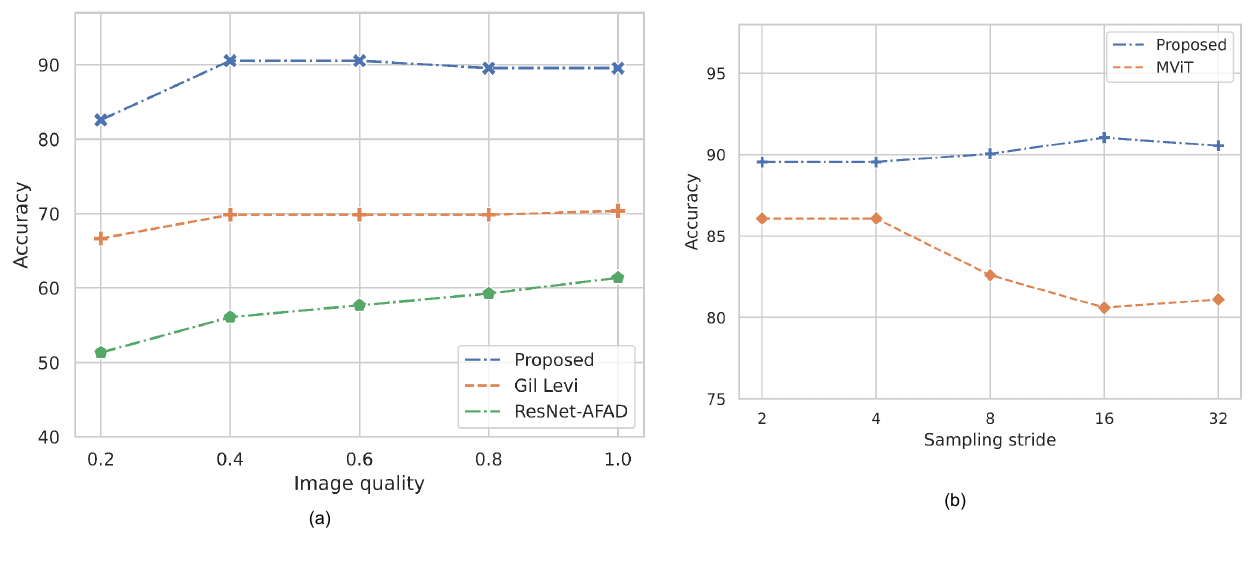}
\end{center}
   \caption{(a) represents the effect of a wide range of input resolution, and (b) shows the different sampling-rates of our proposed method on P-Age dataset compared to other methods.}
\label{fig:res_sampling}
\end{figure*}
\section{Qualitative Analysis}
In this section, we perform two experiments to test the model on i) video quality and ii) variation in sampling-rate. First, we perform experiments to resize the input frames to a low-resolution and bring it back to the input size of the inference $224 \times 224$ to simulate the loss of spatial quality and mimic the low-resolution input. We compare our AgeFormer with other SOTA methods, as shown in Figure \ref{fig:res_sampling}\textcolor{red}{a}. The results in this figure show the robustness of our method against a wide range of low-resolution input that beat existing methods. 

Second, we also experiment our method across various sampling strides to ensure that our model does not rely too much on superficial features, like a sudden change between consecutive frames as an indicator for specific classes, such as kids. Ageformer is mostly stable against these changes, as visualized in Figure \ref{fig:res_sampling}\textcolor{red}{b}.


{\small
\bibliographystyle{ieee_fullname}
\bibliography{egbib}
}